\documentclass{article}

\usepackage{sections/arxiv}

\usepackage[utf8]{inputenc} 
\usepackage[T1]{fontenc}    
\usepackage{hyperref}       
\hypersetup{
    colorlinks=true,
    linkcolor=blue,
    citecolor=blue,
    filecolor=magenta,      
    urlcolor=cyan,
}
\usepackage{url}            
\usepackage{booktabs}       
\usepackage{amsfonts}       
\usepackage{nicefrac}       
\usepackage{microtype}      
\usepackage{lipsum}
\usepackage{helvet}
\usepackage{amsmath}

\usepackage[textwidth=1.8cm,textsize=scriptsize]{todonotes}

\usepackage{soul}

\usepackage{changepage}

\usepackage[utf8]{inputenc}

\usepackage{textcomp,marvosym}

\usepackage{fixltx2e}

\usepackage{amsmath,amssymb}
\usepackage{amsfonts}
\usepackage{amsbsy} 
\usepackage{bbm} 

\usepackage{array,booktabs,calc}

\usepackage{tikz}
\usepackage{standalone}

\usepackage[right]{lineno}

\usepackage{microtype}
\DisableLigatures[f]{encoding = *, family = * }

\usepackage{rotating}

\usepackage{multirow}

\usepackage{pdflscape}

\usepackage{amsmath,amssymb}
\usepackage{amsfonts}
\usepackage{amsbsy} 
\usepackage{bbm} 
\usepackage{mathtools}

\newcommand{\cP}{\mathcal{P}}

\newcommand{\cU}{\mathcal{U}}

\newcommand{\cX}{\mathcal{X}}

\newcommand{\cZ}{\mathcal{Z}}

\newcommand{\D}{\mathbf{D}}

\newcommand{\G}{\mathbf{G}}

\newcommand{\vect}[1]{{\boldsymbol{\mathbf{#1}}}} 
\newcommand{\x}{\vect x}
\newcommand{\y}{\vect y}
\newcommand{\z}{\vect z}

\newcommand{\f}{\vect f}

\newcommand{\kld}{\text{D}_{\text{KL}}}

\newcommand{\remove}[1]{}

\title{Information Theory in Density Destructors}
\author{
  J.~Emmanuel~Johnson\thanks{\url{https://jejjohnson.netlify.app}} \\
  Image Processing Laboratory \\
  Universitat de Val{\`e}ncia\\
  Val{\`e}ncia, Spain\\
  \texttt{juan.johnson@uv.es} \\
  \And
  Valero Laparra\thanks{\url{https://www.uv.es/lapeva/}} \\
  Image Processing Laboratory \\
  Universitat de Val{\`e}ncia\\
  Val{\`e}ncia, Spain\\
  \texttt{valero.laparra@uv.es} \\
  \And
  Gustau Camps-Valls\thanks{\url{https://www.uv.es/gcamps/}} \\
  Image Processing Laboratory \\
  Universitat de Val{\`e}ncia\\
  Val{\`e}ncia, Spain\\
  \texttt{gcamps@uv.es} \\
  \And
  Raul Santos-Rodr\'iguez\thanks{\url{https://www.raulsantosrodriguez.com/}} \\
  Engineering Mathematics Department \\
  University of Bristol\\
  Bristol, UK\\
  \texttt{enrsr@bristol.ac.uk} \\
  \And
  Jesus Malo\thanks{\url{https://isp.uv.es/excathedra.html}} \\
  Image Processing Laboratory \\
  Universitat de Val{\`e}ncia\\
  Val{\`e}ncia, Spain\\
  \texttt{jesus.malo@uv.es} \\
}

\begin{document}
\maketitle

\begin{abstract}
    Density destructors are differentiable and invertible transforms that map multivariate PDFs of arbitrary structure (low entropy) 
into non-structured PDFs (maximum entropy). 
Multivariate Gaussianization and multivariate equalization are specific examples of this family, 
which 
break down the complexity of the original 
PDF through a set of elementary transforms 
that progressively remove the structure of the data.

We demonstrate how this property of density destructive flows is connected to classical information theory, and how density destructors can be used to get more accurate estimates of information theoretic quantities. Experiments with total correlation and mutual information in multivariate sets illustrate the ability of density destructors compared to competing methods.
These results suggest that information theoretic measures may be an alternative optimization criteria when learning density destructive flows.


\end{abstract}

    
\section{Introduction}
Estimating the probability density function (PDF) 
plays a central role in many machine learning problems like regression, classification, or 
data representation.
However, the problem of PDF estimation is notoriously difficult when considering 
moderate and high dimensional data. 
In the deep learning community three families of methods are responsible for the majority of the progress in PDF estimation: 
Variational AutoEncoders (VAEs) \cite{Kingma2014VAES}, Generative Adversarial Networks (GANs) \cite{Goodfellow2014GANS} and Invertible Flows (IFs) \cite{Rezende2015NF}. Each family tackles the PDF estimation from a slightly different {\em algorithmic} perspective, but they share many {\em conceptual} properties. 
They look for two main components: the first looks for a function $\G_\theta (\cdot)$ that maps samples from a known latent space $\cZ$ to the observed space $\cX$. The second component aims to find a function $\D_\theta (\cdot)$ that maps data from our observed space $\cX$ to some latent space $\cZ$. 
Both individual components can be seen, and will be hereafter referred to, as 
density {\em generators} and density {\em destructors}~\cite{Inouye2018DeepDD}. 
The \textbf{generative} transformation can be written as:
%
\begin{equation}
    \z \overset{\G_\theta}{\longrightarrow} \hat \x
\end{equation}
where $\z$ comes from our latent space distribution $\cP_\z$, $\theta$ are the parameters of the generative transformation $\G$, and $\hat \x$ is the approximated data that follows the distribution $\hat \cP_\x$. This component is found in all 
the frameworks mentioned above: the 
generator in GANs, the decoder portion of VAEs, and the invertible function $\f(\cdot)$ in IFs. 
Obtaining this component is difficult as the hypothesis space for $\G_\theta$ is large as we do not know the actual PDF of the data, $\cP_\x$. Thus it is difficult to create appropriate cost functions and clever learning 
schemes are required to obtain
$\G$; i.e. the adversarial formulation in GANs, the encoder-decoder relationship in VAEs, or imposing the invertibility of $\f$ in IFs. 

Alternatively, 
one could look at the problem in the reverse order as a \textbf{destructive} transformation: 
\begin{equation}
    \x \overset{\D_\theta}{\longrightarrow}\hat \z
\end{equation}
%
%
where $\x$ comes from the true data distribution $\cP_\x$, $\theta$ are the parameters of the destructive transformation $\D$, and $\hat \z$ follows the approximated base density $\hat \cP_\z$. This term does not exist in the classical GAN formulation but several new versions have tried to overcome it \cite{InfoGAN, AAE, CycleGAN2017}. In VAEs, this destructor is a non-invertible function (the encoder) where we need to pair it with the decoder $\G$ 
for learning.
In IFs several methods attempt to learn an invertible function $\f$ through inference, mapping the data from $\cX$ to a latent space $\cZ$, \cite{Dinh2017REALNVP, Laparra2011RBIG, Ball2016GDN}. 
Given an invertible transform, $\D$, the relation between our data distribution $\cP_\x$ and the $\cP_\z$ can be calculated through the standard change of variables used in most IFs \cite{Rezende2015NF}:
\begin{equation}
    \cP_\x(\x) = \cP_\z(\hat \z) \left| \nabla_\x\D(\x) \right|\label{eq:cov}
\end{equation}
where $\hat \z=\D(\x)$, $\cP_\z$ is the base distribution, and $\left| \nabla_\x (\cdot) \right|$ is the determinant of the Jacobian of our density destructor $\D$. 
\indent \newline 
The \emph{destructive} perspective gives us some advantages, as we can define a latent distribution $\cP_\z$ with nice enough properties that we can measure how well we approach it.
For example, assuming the base density is uniform, $\cP_\z \sim \cU$, the change-of-variables formula (Eq.~\ref{eq:cov}) just results in the calculation of the exact likelihood of the data because $\cP_\z(\hat \z)$ is equal to one \cite{Inouye2018DeepDD}. Alternatively, we can assume that the base PDF is Gaussian and use standard non-Gaussianity measures 
to assess 
the distance to the goal \cite{Laparra2011RBIG,Ball2016GDN}. In other cases,  sensible cost functions such as the Kullback-Leibler Divergence ($\text{D}_{\text{KL}}$) could be used to measure the similarity between the approximated $\hat{\mathcal{P}}_{\hat z}$ and the true $\mathcal{P}_z$ we choose.


\section{Proposal}
The literature on Invertible Flows \cite{Rezende2015NF,Inouye2018DeepDD} does not link these transforms with classical information theory. In this work we establish this connection by using two properties of density destructive flows: 1)~destructors effectively reduce the data structure so that 
the output may have trivial entropy/redundancy, and 2)~destructors are smooth routes to the target PDF, so their Jacobian can always be computed, which allows us to obtain information measures that ultimately depend on $\nabla_\x\D$. 
Quantifying data structure and the relations between features is at the core of machine learning. Information theoretic magnitudes describe data complexity with few or no assumptions~\cite{Timme2018ATF}. Unfortunately, computation of these magnitudes from their definition is not straightforward because they involve multivariate PDF estimation.
In this work we show how key quantities that describe redundancy, as the Total Correlation,~$T$, \cite{Watanabe1960TC, Studeny1998MI}, and the Mutual Information,~$I$ \cite{Cover06}, naturally appear in the the density destructor framework. Moreover, we will show how, under some conditions, they can be reduced to (easier) univariate operations. 
%
Finally, the experiments demonstrate that information theoretic magnitudes may be effective learning criteria for destructive flows, 
and that estimates of redundancy are obtained via density destructors.

%
%

\section{Information Theory in Density Destructors}
In deep learning, redundancy measures are relevant since they have been linked to the \emph{information bottleneck principle} \cite{Tishby2015INFOBOTTLE} whereby artificial networks can be classified according to the mutual information between layers. Redundancy reduction is also a relevant self-organization principle in natural neural networks \cite{Barlow01,Malo10}, and it is also key in unsupervised learning \cite{Hyvarinen01}.
However, these measures are notoriously difficult to compute in high dimensional data. 

Fortunately, the ability of density destructors to remove  structure makes them appropriate to measure redundancy, as well as to derive convergence rates to the base distribution $\cP_z$ in information terms.

\subsection{Loss Function in density destructors}

The loss function should measure how close the data is to the latent space, $\hat \z$, and follows the base distribution $\cP_\z$. In the latent space we have an advantage because we can choose the target distribution, and typically we choose a distribution such that we have an analytic expression (e.g. Uniform or Gaussian). A usual criterion is to minimize the $\kld$ divergence between the distribution of the transformed data $\hat \cP_\z$ and our target $\cP_\z$ such that: 
\begin{equation}
    J(\hat \z)=\kld \left(\hat \cP_{\hat \z} || \cP_\z \right) 
    \label{eq:kld}
\end{equation}
If the target distribution is separable (just a product of marginals), as usually assumed in destructive flows, we can decompose the above expression as: 
\begin{equation}
    J(\hat \z)= \underbrace{T(\hat \z)}_{\text{Total Corr.}} + \underbrace{J_m(\hat \z)}_{\text{Marginal KLDs}} 
    \label{eq:kld2}
\end{equation}
using the Pythagorean theorem for $\kld$ \cite{Cardoso2003DCGICA}. While the marginal KLDs can be easily reduced by a simple equalization function, in general, the Total Correlation term, $T$, is difficult to compute from its definition since it involves integration of unknown multivariate $\hat \cP_\z$.
However, in order to use the divergence as an optimization criterion, we do not need to compute the value itself; we just have to minimize it; equivalently, we can enforce the difference of $T$ to be maximum before and after the destructor transformation, which is easy to compute as \cite{Studeny1998MI}: 
\vspace{-0.3cm}
\begin{equation}
\Delta T(\x,\hat{\z})=\sum_{d=1}^{D}\left( H (\hat{\z}_d) - H (\x_d) \right) - \mathbb{E}_{\cP_\x}[\log|\nabla_\x\D(\x)|]
\label{stud}
\end{equation}
The first term of the equation is easy to compute since it only involves operations on univariate distributions. The second term is the expected value of the logarithm of the determinant of the Jacobian of the transformation. In the density destructors framework, this transformation is enforced to be smooth and differentiable. Therefore we can compute the second term by evaluating the Jacobian over the training data using automatic differentiation tools. 

\subsection{Estimating information theoretic measures}

In this section, we show how to compute the information theoretic measures $T$ and $I$ (mutual information) following the loss function of density destructors. 
Similar procedures could be used to compute other useful information quantities, such as $\kld$, entropy, and negentropy (non-gaussianity).

\paragraph{Total Correlation.}

$T$ is the information shared among the dimensions of a multidimensional random variable \cite{Watanabe1960TC, Studeny1998MI}.
We are going to show how by applying a density destructor over $\x$ we can compute $T(\x)$ easily as the difference of $T$ between the input and the output, $\Delta T(\x,\hat{\z})$. 
Assuming the density destructor model has reached convergence, the $T$ in the latent space can be computed easily since we know the distribution. Therefore the $T$ of the original data will be the difference in $T$ in $\cX$ plus the $T$ in the latent space $\cZ$, i.e. $T(\x) = \Delta T(\x,\hat{\z}) + T(\hat{\z})$. If the chosen distribution for the latent space is uniform or the Gaussian (as it is customary) 
then the $T$ in the latent space is zero, $T(\hat{\z})=0$. Thus, $T$ of the original data is simply $T(\x) = \Delta T(\x,\hat{\z})$, which could be computed using Eq. \ref{stud}. 
However, note that the expectation over the data set in Eq. \ref{stud} may require many samples and is time consuming.

This inconvenience is solved by the specific density destructor based on Gaussianization proposed in \cite{Laparra2011RBIG}. 
In that case the original PDF is deconstructed through a series of $L$ layers implementing a series of marginal Gaussianization transforms and rotations. Note that both operations in each layer 
are easy to compute (just a set of univariate sigmoids followed by any orthogonal matrix), and they are straightforward to derive and invert. 
In \cite{Laparra2011RBIG} we show the convergence of this procedure to the Gaussian target, but more importantly for the current discussion on $T$, the redundancy of the input is just the sum of the $\Delta T$ in each layer:
\begin{equation}
    T(\x) = \sum_{i = 1}^L\Delta T(\x^i) = \sum_{i=1}^{L} J_{m}(\x^{i+1})
    \label{eq:rbigtc}
\end{equation}
which, as opposed to eq.~\ref{stud}, does not involve any averaging over the whole dataset, and only requires straightforward univariate operations.

\paragraph{Mutual Information.}

$I$ is the amount of information shared by two datasets $\x$ and $\y$ \cite{Cover06}. In the density destructor framework, where $T$ is easy to compute (in general through eq.~\ref{stud}, or in Gaussianization through the simpler eq.~\ref{eq:rbigtc}), $I$ can be computed using \emph{three} density destructors as:
\begin{equation}
I(\x,\y) = T([\D_\x(\x),\D_\y(\y)]).
\label{eq:mirbig}
\end{equation}
where we apply an independent density destructor to each dataset, and then we compute the $T$ for the concatenated variable $[\D_\x(\x), \D_\y(\y)]$ through an extra destructor. 

This procedure is possible because $I$ does not change under invertible transformations (as the density destructors) applied separately to each dataset \cite{Cover06}. Therefore, $I(\x,\y) = I(\D_\x(\x),\D_\y(\y))$. Since we removed $T$ within each individual dataset by applying individual density destructors, the only redundant information that remains in the concatenated vectors is the one shared by both datasets, then $I(\D_\x(\x),\D_\y(\y)) = T([\D_\x(\x),\D_\y(\y)])$. See appendix for more elaborate proof.

\section{Experiments}


For all of our experiments , we assume that the latent space is a Gaussian and our algorithm of choice is the Rotation-Based Iterative Gaussianization (RBIG) \footnote{Please go to the RBIG algorithm homepage for a working implementation along with demonstrations of the IT measures: \url{https://github.com/IPL-UV/rbig}}  \cite{Laparra2011RBIG},  which 
finds a sequence of two steps transformations: univariate Gaussianization procedures coupled with a rotation (e.g. independent components analysis, principal components analysis -PCA- or even random rotations). The two operations (marginal gaussianization and rotation) constitute one layer. We chose PCA for the rotation step in the experiments. 
We use $T$ as an optimization criterion to train the model, and the stopping criterion proposed in \cite{Laparra2011RBIG}. Experiments show that this destructive flow estimates $T$ and $I$ 
effectively compared to other competing algorithms that can be found in the ITE-Toolbox \cite{szabo14itetoolbox}. 

\subsection{Toy Example: Concentric Circles}
\begin{figure}[t]
\begin{center}
\setlength{\tabcolsep}{2pt}
\begin{tabular}{ccc}
(a) $\mathcal X$ & (b) $\hat \z = \D_\theta(\x)$ & (c) $\x = \mathbf D_\theta^{-1} (\hat \z)$\\[3mm]
\hspace{-.5cm}
\includegraphics[width=2.5cm,height=2.5cm]{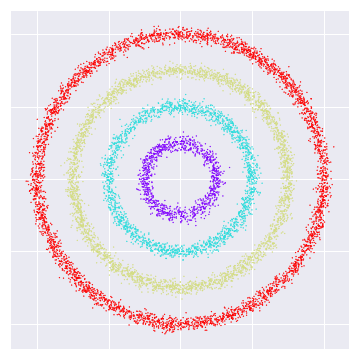} &
\includegraphics[width=2.5cm,height=2.5cm]{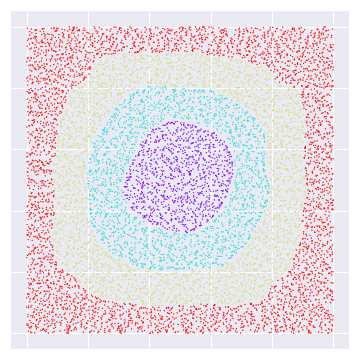} &
\includegraphics[width=2.5cm,height=2.5cm]{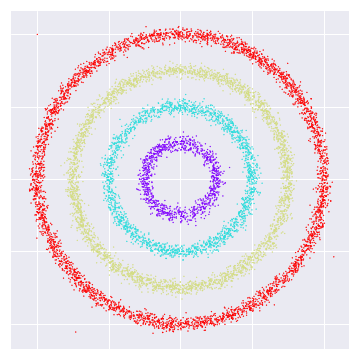} \\
(d) $\mathcal Z$ & (e) $\hat \x = \mathbf D_\theta^{-1} (\z)$ & (f) Cumul. $\Delta$T \\[3mm]
\hspace{-.5cm}
\includegraphics[width=2.7cm,height=2.7cm]{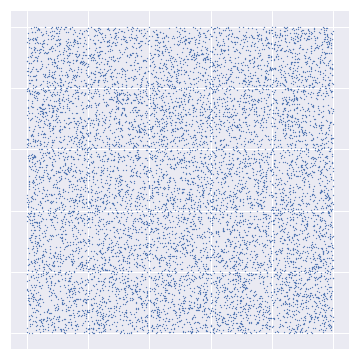} &
\includegraphics[width=2.7cm,height=2.7cm]{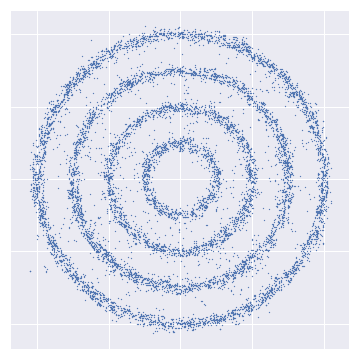} & 
\hspace{-3mm}
\includegraphics[width=2.7cm,height=2.7cm]{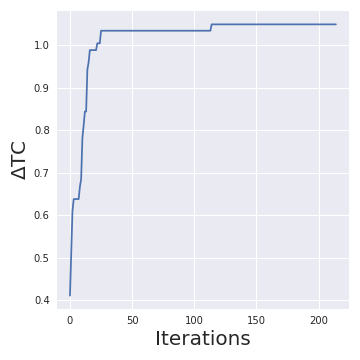}
\\
\end{tabular}
\caption{Density estimation of concentric circles using RBIG: (a) original data distribution $\cX$, (b) approximated base distribution as a unit cube, (c) the inverse of the destructor $\x = \mathbf D_\theta^{-1} (\hat z)$, (d) samples generated from a uniform distribution $\cZ$, (e) the inverse transformation of $\z$ to $\hat \x$, and (f) the Cumulative sum of $\Delta T$ over the layers (or number of iterations). } 
\label{fig:toy}
\end{center}
\end{figure}

We emulated the concentric circles toy example found in \cite{Inouye2018DeepDD}, 
where a multitude of different density destructors that assume a uniform base distribution were used, i.e. a canonical density destructor. The full process can be broken into two parts: 1) minimize the total correlation assuming a Gaussian distribution using RBIG 2) followed by a histogram CDF transformation to project the data into unit hypercube space.
The results shown in fig.~\ref{fig:toy} demonstrate that RBIG is a worthy candidate, and achieves similar results to those in \cite{Inouye2018DeepDD}, both in terms of approximating the data distribution $\cX$ (fig:~\ref{fig:toy} (a-b)) and 
of generating samples from the true base distribution $\z$ (fig:~\ref{fig:toy} (d-e)). We also 
show the quality of the data inversion 
in the approximated base density $\hat z$ (fig:~\ref{fig:toy} (b-c)). 
Figure~\ref{fig:toy}(f) shows the $\Delta$T as the cumulative sum between each layer. Results clearly show that 
we have reached convergence after removing 
{\em all} redundant information. 

\subsection{Total Correlation and Mutual Information}

\begin{figure}[t!]
    \centering
    \begin{tabular}{m{0.1cm} m{3.9cm} m{3.9cm}}
    \hspace{-3.5mm} $d$ &  \hspace{15mm}$\nu$ = 2 &  \hspace{12mm}$\nu$ = 3 \\
    \hspace{-3.5mm} 3 &
    \hspace{-4mm}
    \includegraphics[scale=.26]{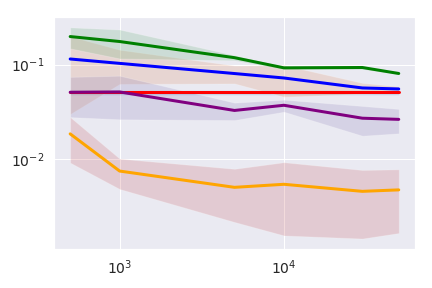} &
    \hspace{-8mm}
    \includegraphics[scale=.26]{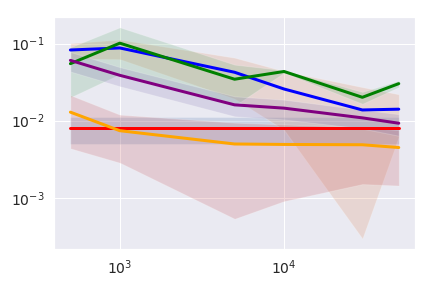} \\[-2.5mm]
    \hspace{-3.5mm}  10 &
    \hspace{-3mm}
    \includegraphics[scale=0.26]{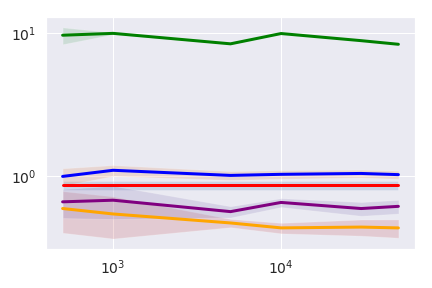} &
    \hspace{-8mm}
    \includegraphics[scale=0.26]{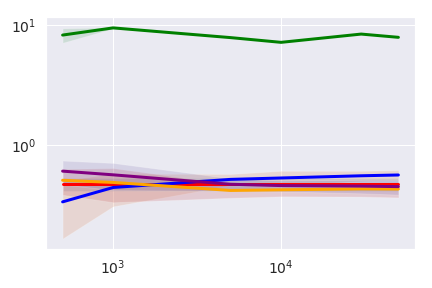} \\[-2.5mm]
    \hspace{-3.5mm}  50 &
    \hspace{-3mm}
    \includegraphics[scale=0.26]{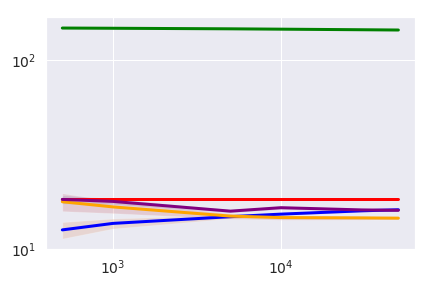} &
    \hspace{-8mm}
    \includegraphics[scale=0.26]{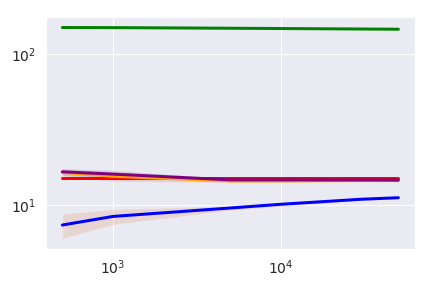}  \\[-1.5mm]
     &  \hspace{15mm}Samples &  \hspace{10mm}Samples
\end{tabular}
    \caption{Estimation of $T$ for data drawn from $d$-dimensional t-Student PDFs with different values of $\nu = 3,5$ and different number of dimensions $d=3,10,50$ respectively. The mean and standard deviation of the results are given for five trials with samples ranging from 500 to 50,000. \textbf{Legend}: Analytical (red), RBIG (purple), expF (orange), and vME (green).}
    \label{fig:TC_stud}
\end{figure}

\begin{figure}[t!]
    \centering
    \begin{tabular}{m{0.1cm} m{3.9cm} m{3.9cm}}
    \hspace{-3.5mm} $d$ &  \hspace{15mm}$\nu$ = 2 &  \hspace{12mm}$\nu$ = 3 \\
    \hspace{-3.5mm} 3 &
    \hspace{-4mm}
    \includegraphics[scale=.26]{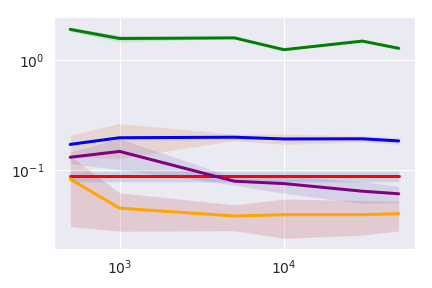} &
    \hspace{-8mm}
    \includegraphics[scale=.26]{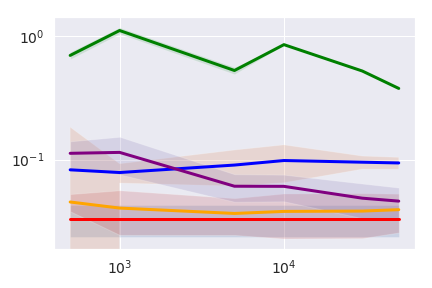} \\[-2.5mm]
    \hspace{-3.5mm}  10 &
    \hspace{-3mm}
    \includegraphics[scale=0.26]{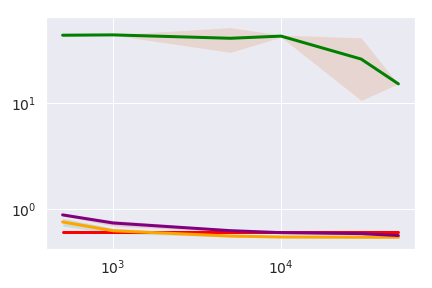} &
    \hspace{-8mm}
    \includegraphics[scale=0.26]{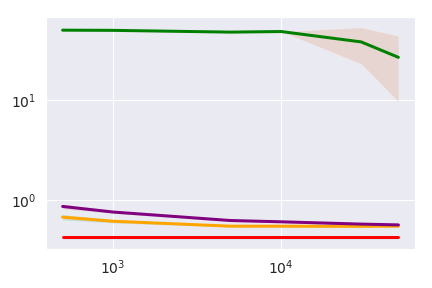} \\[-2.5mm]
    \hspace{-3.5mm}  50 &
    \hspace{-3mm}
    \includegraphics[scale=0.26]{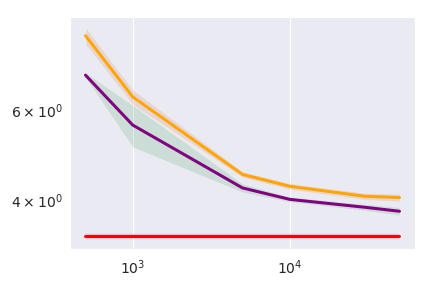} &
    \hspace{-8mm}
    \includegraphics[scale=0.26]{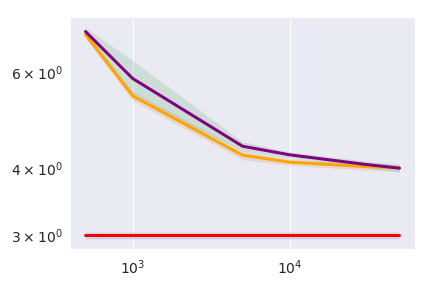} \\[-1.5mm]
     &  \hspace{15mm}Samples &  \hspace{10mm}Samples
\end{tabular}
    \caption{Estimation of $I$ for data drawn from $d$-dimensional t-Student PDFs with different values of $\nu = 3,5$ and different number of dimensions $d=3,10,50$ respectively. The mean and standard deviation of the results are given for five trials with samples ranging from 500 to 50,000. \textbf{Legend}: Analytical (red), RBIG (purple), expF (orange), and vME (green).}
    \label{fig:MI_stud}
\end{figure}

We used the RBIG destructive flow to measure the $T$ and $I$ found within data drawn from multivariate t-Student distributions.
Our redundancy estimates are compared with 
the values found using the k-Nearest Neighbor (kNN) \cite{Goria2005ITkNN}, 
the maximum likelihood expectation with the analytical value of the exponential family (expF) \cite{Nielsen2010expF}, and the von Mises Expansion (vME) \cite{Kandasamy2015vonmises} in the implementations given in the ITE-Toolbox \cite{szabo14itetoolbox}. 
A comparison in Fig.~\ref{fig:TC_stud} and Fig.~\ref{fig:MI_stud} are done in terms of distance to the analytical values for $T$ and $I$ in the t-Student \cite{Guerrero-Cusumano1998}. Any algorithms omitted from the plots resulted in negative values for the respective IT measures. Results for the RBIG destructive flow (in purple) are always the best or close to the best, showing that it is a robust method to compute multivariate information theoretic measures.


\section{Conclusion}

We connected  
the density destructors framework introduced in \cite{Inouye2018DeepDD} with classical information theory. 
This connection allows the use of Total Correlation as learning criterion for destructive 
flows and to compute non-trivial information theoretic quantities via density destructors. 
We chose a particular density destructive flow for multivariate Gaussianization and reported empirical evidence of performance in simulated examples. 

\newpage

\section*{Acknowledgements}
This work was funded by MINECO: DPI2017-89867.

\bibliographystyle{apalike} 
\bibliography{bibs/main}

\section{Appendix: Mutual Information from density destructors}


Recall the definitions for I and T as:
\begin{eqnarray*}
    I(\x,\y) &=&  H(\x) + H(\y) - H([\x,\y]) \\
    T(\x) &=& \sum_{d=1}^D H(\x_d) - H(\x)
\end{eqnarray*}
If we apply two separate density destructor transforms on $\x \in \mathbb{R}^{D_x}$ and $\y \in \mathbb{R}^{D_x}$, we achieve the new datasets $\hat \x$ and $\hat \y$ respectively, where $\sum_{d=1}^{D_x} H(\hat\x_d) = H(\hat\x)$ and $\sum_{d=1}^{D_y} H(\hat\y_d) = H(\hat\y)$ . We can rewrite the mutual information in terms of the transformed versions like so:
\begin{eqnarray*}
    I(\hat\x, \hat\y) &=& \sum_{d=1}^{D_x} H(\hat\x_d) + \sum_{d=1}^{D_y} H(\hat\y_d) - H([\hat\x, \hat\y])
\end{eqnarray*}
For convenience lets assume that we stack $\hat\x$ and $\hat\y$ into a single vector $\hat{{\vect v}} = [\hat\x, \hat\y]$, then we can combine the summations for the marginals into a single term that runs through all the dimensions of $\hat{{\vect v}}$:
\begin{eqnarray*}
    I(\hat\x, \hat\y) &=& \sum_{d=1}^{D_x + D_y} H(\hat{\vect v}_d) - H({\vect v}) 
    \end{eqnarray*}
    And then applying the definition of total correlation:
    \begin{eqnarray*}
    I(\hat\x, \hat\y) &=& T({\vect v}) = T([\hat\x,\hat\y]),
\end{eqnarray*}
leading to eq.~\ref{eq:mirbig}.

So we see that the mutual information for two destructed variables is the same as the total correlation of the two destructed variables stacked into a single vector. The mutual information is invariant under smooth, invertible transformations, as is the case for any density destructors applied to $\x$ and $\y$. The role of these initial destructors is removing redundant information between the different variables within each dataset. Once we did that, the remaining redundancy (in the stacked vector, which will be computed by the third destructor) is the information shared by the original variables. If this third destructor is chosen to be the Rotation-Based Iterative Gaussianization \cite{Laparra2011RBIG}, we have an easy way, eq.\ref{eq:rbigtc}, to calculate the mutual information between two multivariate variables of arbitrary dimension.

\end{document}